# An Empirical Analysis of Search in GSAT

**Ian P. Gent** I.P.GENT@EDINBURGH.AC.UK
*Department of Artificial Intelligence, University of Edinburgh*
*80 South Bridge, Edinburgh EH1 1HN, United Kingdom*

**Toby Walsh** WALSH@LORIA.FR
*INRIA-Lorraine, 615, rue du Jardin Botanique,*
*54602 Villers-les-Nancy, France*

## Abstract

We describe an extensive study of search in GSAT, an approximation procedure for propositional satisfiability. GSAT performs greedy hill-climbing on the number of satisfied clauses in a truth assignment. Our experiments provide a more complete picture of GSAT's search than previous accounts. We describe in detail the two phases of search: rapid hill-climbing followed by a long plateau search. We demonstrate that when applied to randomly generated 3-SAT problems, there is a very simple scaling with problem size for both the mean number of satisfied clauses and the mean branching rate. Our results allow us to make detailed numerical conjectures about the length of the hill-climbing phase, the average gradient of this phase, and to conjecture that both the average score and average branching rate decay exponentially during plateau search. We end by showing how these results can be used to direct future theoretical analysis. This work provides a case study of how computer experiments can be used to improve understanding of the theoretical properties of algorithms.

## 1. Introduction

Mathematicians are increasingly recognizing the usefulness of experiments with computers to help advance mathematical theory. It is surprising therefore that one area of mathematics which has benefitted little from empirical results is the theory of algorithms, especially those used in AI. Since the objects of this theory are abstract descriptions of computer programs, we should in principle be able to reason about programs entirely deductively. However, such theoretical analysis is often too complex for our current mathematical tools. Where theoretical analysis is practical, it is often limited to (unrealistically) simple cases. For example, results presented in (Koutsoupias & Papadimitriou, 1992) for the greedy algorithm for satisfiability do not apply to interesting and hard region of problems as described in §3. In addition, actual behaviour on real problems is sometimes quite different to worst and average case analyses. We therefore support the calls of McGeoch (McGeoch, 1986), Hooker (Hooker, 1993) and others for the development of an empirical science of algorithms. In such a science, experiments as well as theory are used to advance our understanding of the properties of algorithms. One of the aims of this paper is to demonstrate the benefits of such an empirical approach. We will present some surprising experimental results and demonstrate how such results can direct future efforts for a theoretical analysis.

The algorithm studied in this paper is GSAT, a randomized hill-climbing procedure for propositional satisfiability (or SAT) (Selman, Levesque, & Mitchell, 1992; Selman & Kautz, 1993a). Propositional satisfiability is the problem of deciding if there is an assignment for the variables in a propositional formula that makes the formula true. Recently, there has been considerable interest in GSAT as it appears to be able to solve large and difficult satisfiability problems beyond the range of conventional procedures like Davis-Putnam (Selman et al., 1992). We believe that the results we give here will actually apply to a larger family of procedures for satisfiability called GENSAT (Gent & Walsh, 1993). Understanding such procedures more fully is of considerable practical interest since SAT is, in many ways, the archetypical (and intractable) NP-hard problem. In addition, many AI problems can be encoded quite naturally in SAT (*eg.* constraint satisfaction, diagnosis and vision interpretation, refutational theorem proving, planning).

This paper is structured as follows. In §2 we introduce GSAT, the algorithm studied in the rest of the paper. In §3 we define and motivate the choice of problems used in our experiments. The experiments themselves are described in §4. These experiments provide a more complete picture of GSAT's search than previous informal accounts. The results of these experiments are analysed more closely in §5 using some powerful statistical tools. This analysis allow us to make various experimentally verifiable conjectures about GSAT's search. For example, we are able to conjecture: the length of GSAT's initial hill-climbing phase; the average gradient of this phase; the simple scaling of various important features like the score (on which hill-climbing is performed) and the branching rate. In §6 we suggest how such results can be used to direct future theoretical analysis. Finally, in §7 we describe related work and end with some brief conclusions in §8.

## 2. GSAT

GSAT is a random greedy hill-climbing procedure. GSAT deals with formulae in conjunctive normal form (CNF); a formula, $\Sigma$ is in CNF iff it is a conjunction of clauses, where a clause is a disjunction of literals. GSAT starts with a randomly generated truth assignment, then hill-climbs by "flipping" the variable assignment which gives the largest increase in the number of clauses satisfied (called the "score" from now on). Given the choice between several equally good flips, GSAT picks one at random. If no flip can increase the score, then a variable is flipped which does not change the score or (failing that) which decreases the score the least. Thus GSAT starts in a random part of the search space and searches for a global solution using only local information. Despite its simplicity, this procedure has been shown to give good performance on hard satisfiability problems (Selman et al., 1992).

```
procedure GSAT(Σ)
  for i := 1 to Max-tries
    T := random truth assignment
    for j := 1 to Max-flips
      if T satisfies Σ then return T
      else Poss-flips := set of vars which increase satisfiability most
           V := a random element of Poss-flips
           T := T with V's truth assignment flipped
    end
  end
  return "no satisfying assignment found"
```

In (Gent & Walsh, 1993) we describe a large number of experiments which suggest that neither greediness not randomness is important for the performance of this procedure. These experiments also suggest various other conjectures. For instance, for random 3-SAT problems (see §3) the log of the runtime appears to scale with a less than linear dependency on the problem size. Conjectures such as these could, as we noted in the introduction, be very profitably used to direct future efforts to analyse GSAT theoretically. Indeed, we believe that the experiments reported here suggest various conjectures which would be useful in a proof of the relationship between runtime and problem size (see §6 for more details)

## 3. Problem Space

To be able to perform experiments on an algorithm, you need a source of problems on which to run the algorithm. Ideally the problems should come from a probability distribution with some well-defined properties, contain a few simple parameters and be representative of problems which occur in real situations. Unfortunately, it is often difficult to meet all these criteria. In practice, one is usually forced to accept either problems from a well-defined distribution with a few simple parameters or a benchmark set of real problems, necessarily from some unknown distribution. In these experiments we adopt the former approach and use CNF formulae randomly generated according to the random $k$-SAT model.

Problems in random $k$-SAT with N variables and L clauses are generated as follows: a random subset of size $k$ of the N variables is selected for each clause, and each variable is made positive or negative with probability $\frac{1}{2}$. For random 3-SAT, there is a phase transition from satisfiable to unsatisfiable when L is approximately 4.3N (Mitchell, Selman, & Levesque, 1992; Larrabee & Tsuji, 1992; Crawford & Auton, 1993). At lower L, most problems generated are under-constrained and are thus satisfiable; at higher L, most problems generated are over-constrained and are thus unsatisfiable. As with many NP-complete problems, problems in the phase transition are typically much more difficult to solve than problems away from the transition (Cheeseman, Kanefsky, & Taylor, 1991). The region L=4.3N is thus generally considered to be a good source of hard SAT problems and has been the focus of much recent experimental effort.

## 4. GSAT's search

When GSAT was first introduced, it was noted that search in each try is divided into two phases. In the first phase of a try, each flip increases the score. However, this phase is relatively short and is followed by a second phase in which most flips do not increase the score, but are instead sideways moves which leave the same number of clauses satisfied. This phase is a search of a "plateau" for the occasional flip that can increase the score.[1] One of the aims of this paper is to improve upon such informal observations by making *quantitative* measurements of GSAT's search, and by using these measurements to make several experimentally testable predictions.

1. Informal observations to this effect were made by Bart Selman during the presentation of (Selman et al., 1992) at AAAI-92. These observations were enlarged upon in (Gent & Walsh, 1992).

In our experiments, we followed three methodological principles from (McGeoch, 1986). First, we performed experiments with large problem sizes and many repetitions, to reduce variance and allow for emergent properties. Second, we sought good views of the data. That is, we looked for features of performance which are meaningful and which are as predictable as possible. Third, we analysed our results closely. Suitable analysis of data may show features which are not clear from a simple presentation. In the rest of this paper we show how these principles enabled us to make very detailed conjectures about GSAT's search.

Many features of GSAT's search space can be graphically illustrated by plotting how they vary during a try. The most obvious feature to plot is the score, the number of satisfied clauses. In our quest for a good view of GSAT's search space, we also decided to plot "poss-flips" at each flip: that is, the number of equally good flips between which GSAT randomly picks. This is an interesting measure since it indicates the branching rate of GSAT's search space.

We begin with one try of GSAT on a 500 variable random 3-SAT problem in the difficult region of L = 4.3N (Figure 1a). Although there is considerable variation between tries, this graph illustrates features common to all tries. Both score (in Figure 1a) and poss-flips (in Figure 1b) are plotted as percentages of their maximal values, that is L and N respectively. The percentage score starts just above 87.5%, which might seem surprisingly high. Theoretically, however, we expect a random truth assignment in $k$-SAT to satisfy $\frac{2^k-1}{2^k}$ of all clauses (in this instance, $\frac{7}{8}$). As expected from the earlier informal description, the score climbs rapidly at first, and then flattens off as we mount the plateau. The graph is discrete since positive moves increase the score by a fixed amount, but some of this discreteness is lost due to the small scale. To illustrate the discreteness, in Figure 1b we plot the change in the number of satisfied clauses made by each flip (as its exact value, unscaled). Note that the $x$-axis for both plots in Figure 1b is the same.

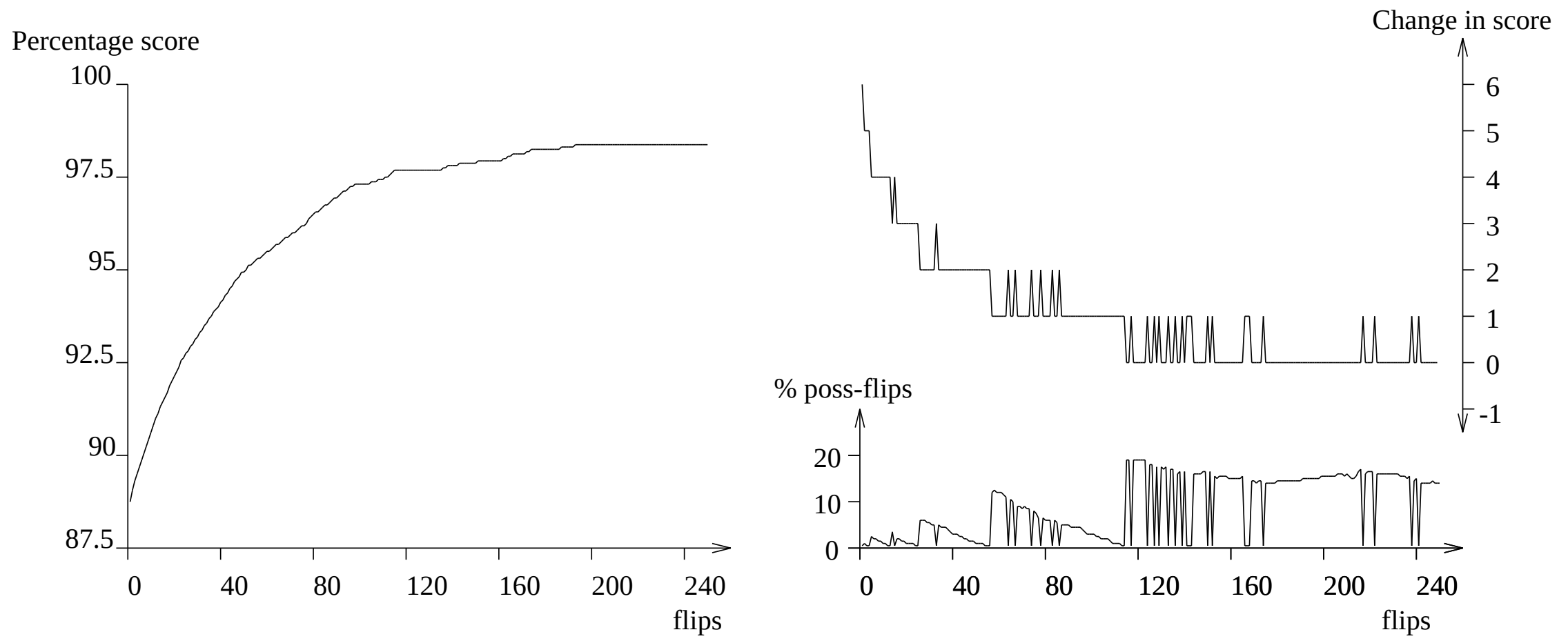

(a) Score (b) Change in score, and poss-flips

Figure 1: GSAT's behaviour during one try, N = 500, L = 2150, first 250 flips

The behaviour of poss-flips is considerably more complicated than that of the score. It is easiest first to consider poss-flips once on the plateau. The start of plateau search, after 115 flips, coincides with a very large increase in poss-flips, corresponding to a change from the region where a small number of flips can increase the score by 1 to a region where a large number of flips can be made which leave the score unchanged. Once on the plateau, there are several sharp dips in poss-flips. These correspond to flips where an increase by 1 in the score was effected, as can be seen from Figure 1b. It seems that if you can increase the score on the plateau, you only have a very small number of ways to do it. Also, the dominance of flips which make no change in score graphically illustrates the need for such "sideways" flips, a need that has been noted before (Selman et al., 1992; Gent & Walsh, 1993).

Perhaps the most fascinating feature is the initial behaviour of poss-flips. There are four well defined wedges starting at 5, 16, 26, and 57 flips, with occasional sharp dips. These wedges demonstrate behaviour analogous to the that of poss-flips on the plateau. The plateau spans the region where flips typically do not change the score: we call this region $H_0$ since hill-climbing typically makes zero change to the score. The last wedge spans the region $H_1$ where hill-climbing typically increases the score by 1, as can be seen very clearly from Figure 1b. Again Figure 1b shows that the next three wedges (reading right to left) span regions $H_2$, $H_3$, and $H_4$. As with the transition onto the plateau, the transition between each region is marked by a sharp increase in poss-flips. Dips in the wedges represent unusual flips which increase the score by more than the characteristic value for that region, just as the dips in poss-flips on the plateau represent flips where an increase in score was possible. This exact correlation can be seen clearly in Figure 1b. Note that in this experiment, in no region $H_j$ did a change in score of $j+2$ occur, and that there was no change in score of $-1$ at all. In addition, each wedge in poss-flips appears to decay close to linearly. This is explained by the facts that once a variable is flipped it no longer appears in poss-flips (flipping it back would decrease score), that most of the variables in poss-flips can be flipped independently of each other, and that new variables are rarely added to poss-flips as a consequence of an earlier flip. On the plateau, however, when a variable is flipped which does not change the score, it remains in poss-flips since flipping it back also does not change the score.

To determine if this behaviour is typical, we generated 500 random 3-SAT problems with N=500 and L=4.3N, and ran 10 tries of GSAT on each problem. Figure 2a shows the mean percentage score[2] while Figure 2b presents the mean percentage poss-flips together with the mean change in score at each flip. (The small discreteness in this figure is due to the discreteness of Postscript's plotting.)

The average percentage score is very similar to the behaviour on the individual run of Figure 1, naturally being somewhat smoother. The graph of average poss-flips seems quite different, but it is to be expected that you will neither observe the sharply defined dips in poss-flips from Figure 1b, nor the very sharply defined start to the wedges, since these happen at varying times. It is remarkable that the wedges are consistent enough to be visible when averaged over 5,000 tries; the smoothing in the wedges and the start of the plateau is caused by the regions not starting at exactly the same time in each try.

Figure 2 does not distinguish between satisfiable and unsatisfiable problems. There is no current technique for determining the satisfiability of 500 variable 3-SAT problems in feasible time. From instances we have been able to test, we do not believe that large

2. In this paper we assign a score of 100% to flips which were not performed because a satisfying truth assignment had already been found.

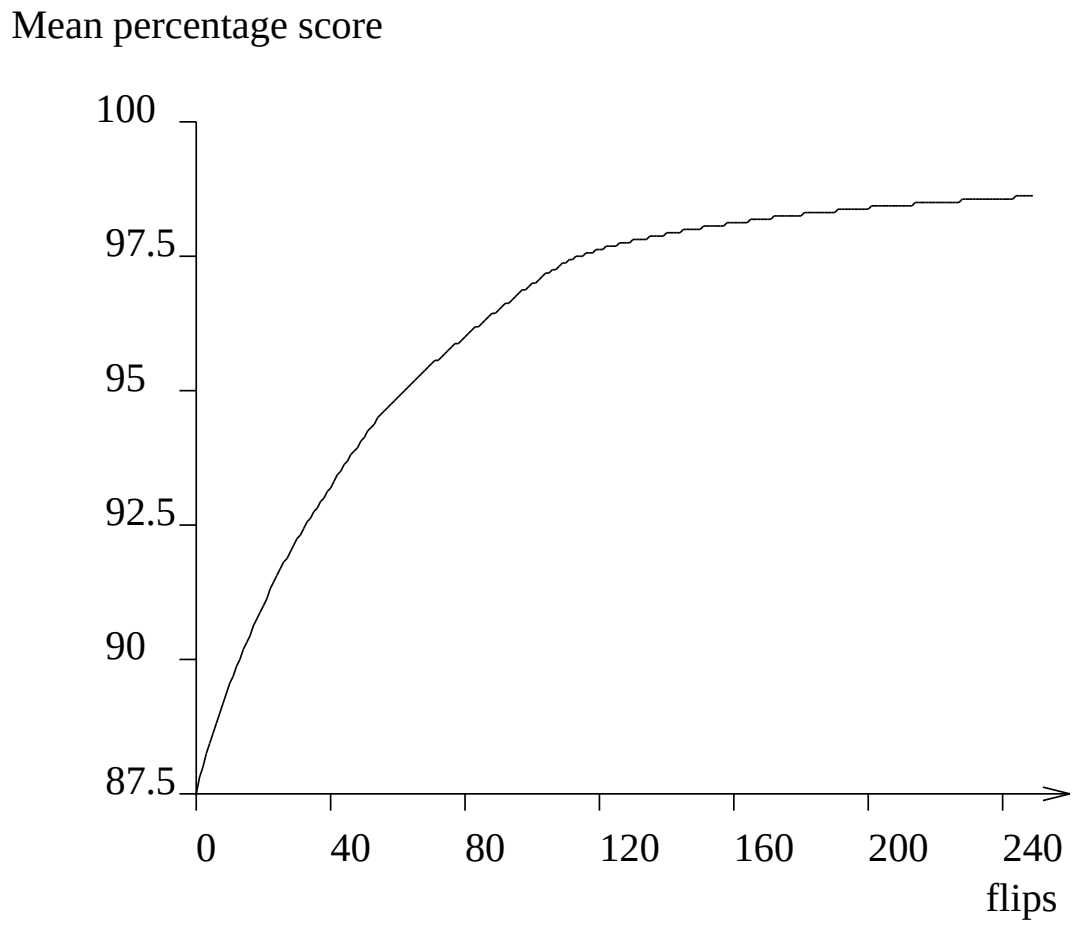

(a) Mean score

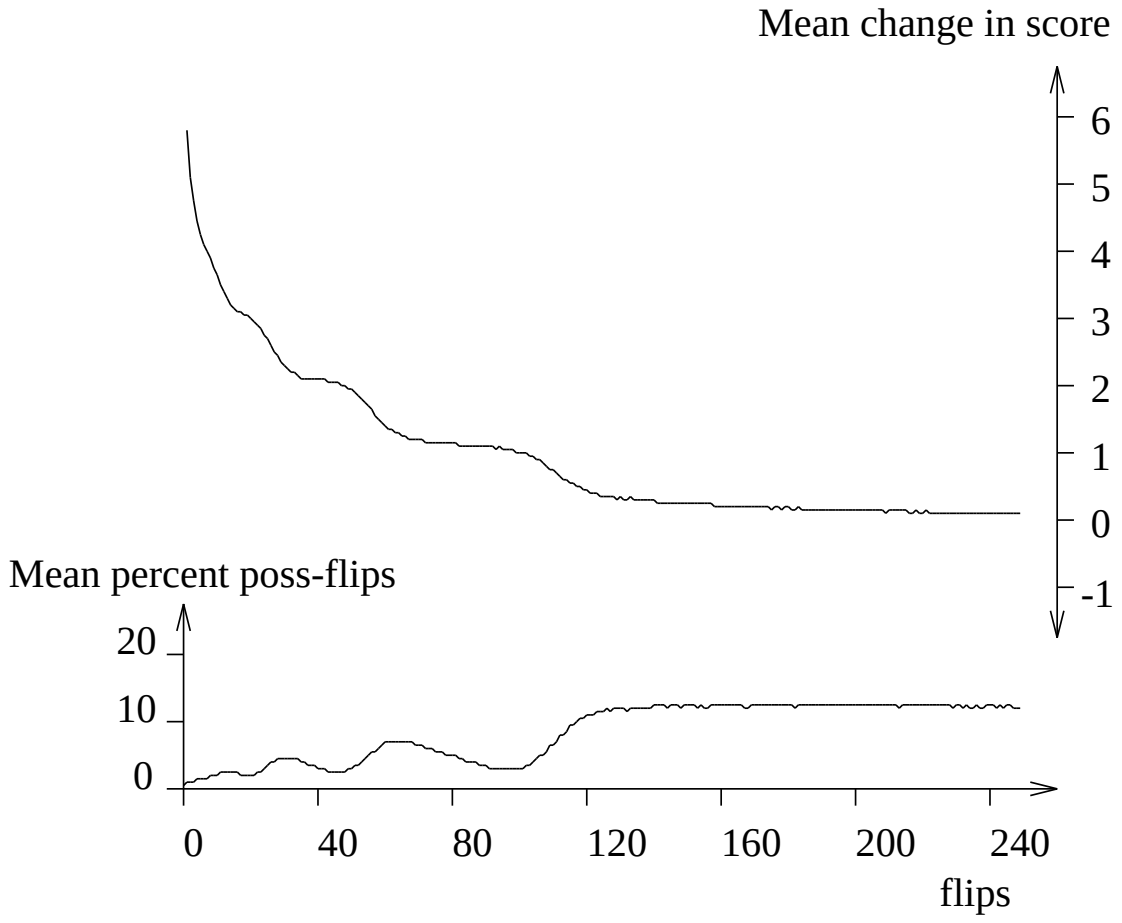

(b) Mean change in score, poss-flips

Figure 2: Mean GSAT behaviour, N = 500, L = 4.3N, first 250 flips

differences from Figure 2 will be seen when it is possible to plot satisfiable and unsatisfiable problems separately, but this remains an interesting topic to investigate in the future.

Experiments with other values of N with the same ratio of clauses to variables demonstrated qualitatively similar behaviour. More careful analysis shows the remarkable fact that not only is the behaviour qualitatively similar, but quantitatively similar, with a simple linear dependency on N. If graphs similar to Figure 2 are plotted for each N with the x-axis scaled by N, behaviour is almost *identical.* To illustrate this, Figure 3 shows the mean percentage score, percentage poss-flips, and change in score, for N = 500, 750, and 1000, for L = 4.3N and for the first 0.5N flips (250 flips at N = 500). Both Figure 3a and Figure 3b demonstrate the closeness of the scaling, to the extent that they may appear to contain just one thick line. In Figure 3b there is a slight tendency for the different regions of hill-climbing to become better defined with increasing N.

The figures we have presented only reach a very early stage of plateau search. To investigate further along the plateau, we performed experiments with 100, 200, 300, 400, and 500 variables from 0 to 2.5N flips.[3] In Figure 4a shows the mean percentage score in each case, while Figure 4b shows the mean percentage poss-flips, magnified on the $y$-axis for clarity. Both these figures demonstrate the closeness of the scaling on the plateau. In Figure 4b the graphs are not quite so close together as in Figure 4a. The phases of hill-climbing become much better defined with increasing N. During plateau search, although separate lines are distinguishable, the difference is always considerably less than 1% of the total number of variables.

The problems used in these experiments (random 3-SAT with L=4.3N) are believed to be unusually hard and are satisfiable with probability approximately $\frac{1}{2}$. Neither of these facts appears to be relevant to the scaling of GSAT's search. To check this we performed a similar range of experiments with a ratio of clauses to variables of 6. Although almost all such problems are unsatisfiable, we observed exactly the same scaling behaviour. The score

3. At 100 variables, 2.5N flips is close to the optimal value for Max-flips. However, experiments have suggested that Max-flips may need to vary quadratically for larger N (Gent & Walsh, 1993).

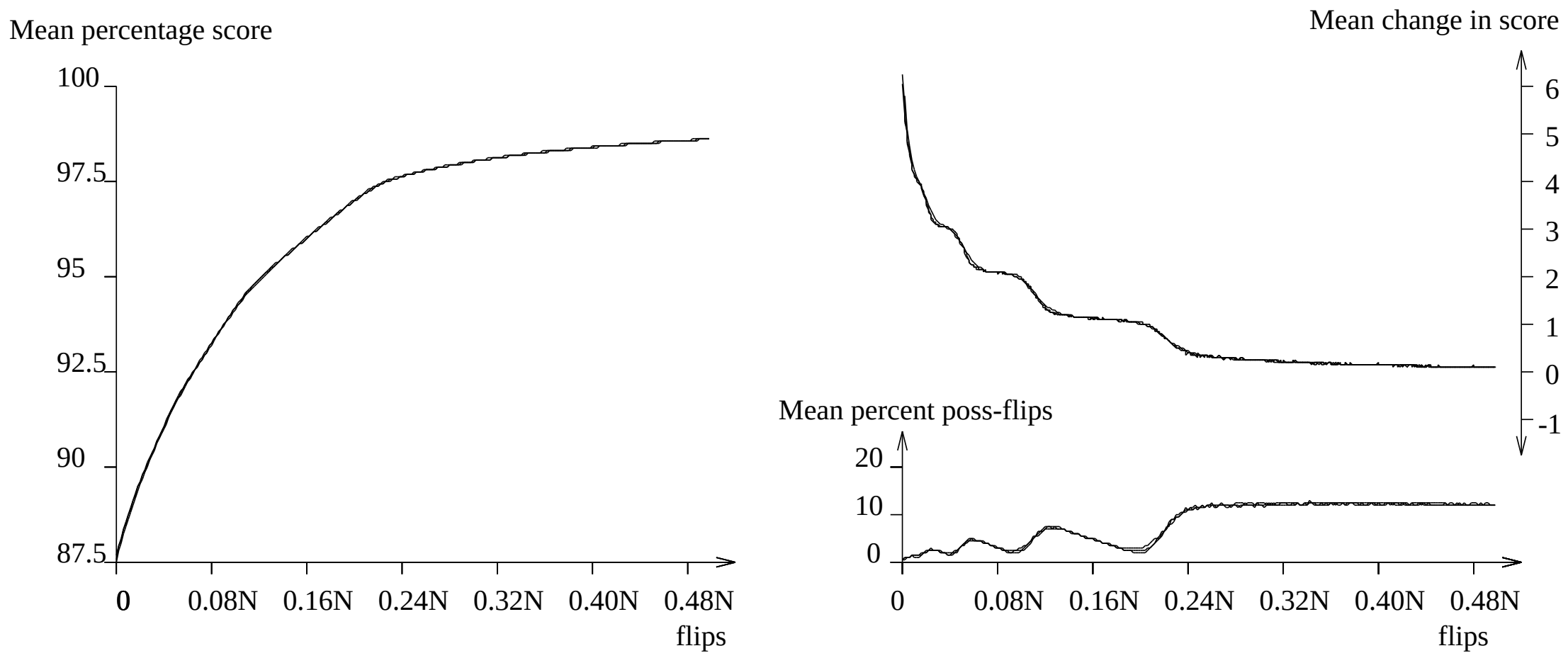

(a) Mean score (b) Mean change in score, poss-flips

Figure 3: Scaling of mean GSAT behaviour, N = 500, 750, 1000, first 0.5N flips

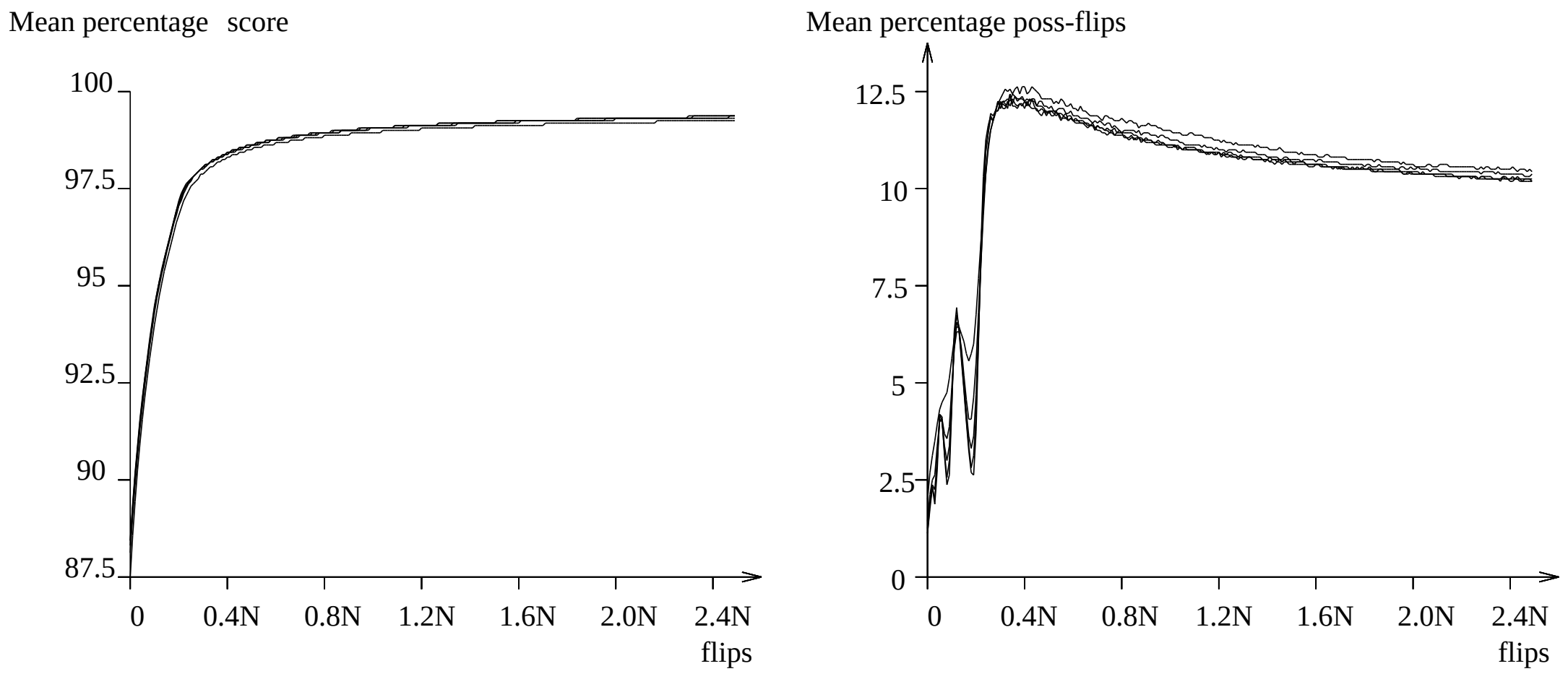

(a) mean score, L = 4.3N (b) mean poss-flips, L = 4.3N

Figure 4: Scaling of mean GSAT behaviour, N = 100, 200, 300, 400, 500

does not reach such a high value as in Figure 4a, as is to be expected, but nevertheless shows the same linear scaling. On the plateau, the mean value of poss-flips is lower than before. We again observed this behaviour for L = 3N, where almost all problems are satisfiable. The score approaches 100% faster than before, and a higher value of poss-flips is reached on the plateau, but the decay in the value of poss-flips seen in Figure 4b does not seem to be present.

To summarise, we have shown that GSAT's hill-climbing goes through several distinct phases, and that the average behaviour of certain important features scale in linear fashion with N. These results provide a considerable advance on previous informal descriptions of GSAT's search.

## 5. Numerical Conjectures

In this section, we will show that detailed numerical conjectures can be made if the data presented graphically in §4 is analysed numerically. We divide our analysis into two parts: first we deal with the plateau search, where behaviour is relatively simple, then we analyse the hill-climbing search.

On the plateau, both average score and poss-flips seem to decay exponentially with a simple linear dependency on problem size. To test this, we performed regression analysis on our experimental data, using the models

$$S(x) = \mathrm{N} \cdot (B - C \cdot e^{-\frac{x}{A \cdot \mathrm{N}}}) \quad (1)$$
$$P(x) = \mathrm{N} \cdot (E + F \cdot e^{-\frac{x}{D \cdot \mathrm{N}}}) \quad (2)$$

where $x$ represents the number of flips, $S(x)$ the average score at flip $x$ and $P(x)$ the average number of possible flips. To determine GSAT's behaviour just on the plateau, we analysed data on mean score, starting from 0.4N flips, a time when plateau search always appears to have started (see §5). Our experimental data fitted the model very well. Detailed results for N = 500 are given in Table 1 to three significant figures. The values of A, B, and C change only slightly with N, providing further evidence for the scaling of GSAT's behaviour. For L = 3N the asymptotic mean percentage score is very close to 100% of clauses being satisfied, while for L = 4.3N it is approximately 99.3% of clauses and for L = 6N it is approximately 98.2% of clauses. A good fit was also found for mean poss-flips behaviour (see Table 2 for N = 500), except for L = 3N, where the mean value of poss-flips on the plateau may be constant. It seems that for L = 4.3N the asymptotic value of poss-flips is about 10% of N and that for 6 it is about 5% of N.

It is important to note that the behaviour we analysed was for mean behaviour over both satisfiable and unsatisfiable problems. It is likely that individual problems will exhibit similar behaviour with different asymptotes, but we do not expect even satisfiable problems to yield a mean score of 100% asymptotically. Note that as N increases a small error in percentage terms may correspond to a large error in the actual score. As a result, our predictions of asymptotic score may be inaccurate for large N, or for very large numbers of flips. Further experimentation is necessary to examine these issues in detail.

| L/N | N | A | B | C | $R^2$ |
|---|---|---|---|---|---|
| 3 | 500 | 0.511 | 2.997 | 0.0428 | 0.995 |
| 4.3 | 500 | 0.566 | 4.27 | 0.0772 | 0.995 |
| 6 | 500 | 0.492 | 5.89 | 0.112 | 0.993 |

Table 1: Regression results for average score of GSAT.[4]

4. The value of $R^2$ is a number in the interval $[0,1]$ indicating how well the variance in data is explained by the regression formula. $1 - R^2$ is the ratio between variance of the data from its predicted value, and the variance of the data from the mean of all the data. A value of $R^2$ close to 1 indicates that the regression formula fits the data very well.

| L/N | N | D | E | F | $R^2$ |
|---|---|---|---|---|---|
| 4.3 | 500 | 0.838 | 0.100 | 0.0348 | 0.996 |
| 6 | 500 | 0.789 | 0.0502 | 0.0373 | 0.999 |

Table 2: Regression results on average poss-flips of GSAT.

We have also analysed GSAT's behaviour during its hill-climbing phase. Figure 1b shows regions where most flips increase the score by 4, then by 3, then by 2, then by 1. Analysis of our data suggested that each phase lasts roughly twice the length of the previous one. This motivates the following conjectures: GSAT moves through a sequence of regions $H_j$ for $j = ..., 3, 2, 1$ in which the majority of flips increase the score by $j$, and where the length of each region $H_j$ is proportional to $2^{-j}$ (except for the region $H_0$ which represents plateau search).

To investigate this conjecture, we analysed 50 tries each on 20 different problems for random 3-SAT problems at N=500 and L=4.3N. We very rarely observe flips in $H_j$ that increase the score by *less* than $j$, and so define $H_j$ as the region between the first flip which increases the score by exactly $j$ and the first flip which increases the score by less than $j$ (unless the latter actually appears before the former, in which case $H_j$ is empty). One simple test of our conjecture is to compare the total time spent in $H_j$ with the total time up to the end of $H_j$; we predict that this ratio will be $\frac{1}{2}$. For $j = 1$ to 4 the mean and standard deviations of this ratio, and the length of each region are shown in Table 3.[5] This data supports our conjecture although as j increases each region is slightly longer than predicted. The total length of hill-climbing at N=500 is 0.22N flips, while at N=100 it is 0.23N. This is consistent with the scaling behaviour observed in §4.

| Region | mean ratio | s.d. | mean length | s.d. |
|---|---|---|---|---|
| All climbing | — | — | 112 | 7.59 |
| $H_1$ | 0.486 | 0.0510 | 54.7 | 7.69 |
| $H_2$ | 0.513 | 0.0672 | 29.5 | 5.12 |
| $H_3$ | 0.564 | 0.0959 | 15.7 | 3.61 |
| $H_4$ | 0.574 | 0.0161 | 7.00 | 2.48 |

Table 3: Comparative and Absolute Lengths of hill-climbing phases

Our conjecture has an appealing corollary. Namely, that if there are $i$ non-empty hill-climbing regions, the average change in score per flip during hill-climbing is:

$$\frac{1}{2} \cdot 1 + \frac{1}{4} \cdot 2 + \frac{1}{8} \cdot 3 + \frac{1}{16} \cdot 4 + \cdots + \frac{1}{2^i} \cdot i \quad \approx \quad 2. \qquad (3)$$

It follows from this that mean gradient of the entire hill-climbing phase is approximately 2. At N=500, we observed a mean ratio of change in score per flip during hill-climbing of 1.94

5. The data for "All climbing" is the length to the start of $H_0$.

with a standard deviation of 0.1. At N=100, the ratio is 1.95 with a standard deviation of 0.2.

The model presented above ignores flips in $H_j$ which increase the score by more than $j$. Such flips were seen in Figure 1b in regions $H_3$ to $H_1$. In our experiment 9.8% of flips in $H_1$ were of size 2 and 6.3% of flips in $H_2$ were of size 3. However, flips of size $j+2$ were very rare, forming only about 0.02% of all flips in $H_1$ and $H_2$. We conjectured that an exponential decay similar to that in $H_0$ occurs in each $H_j$. That is, we conjecture that the average change in number of satisfied clauses from flip $x$ to flip $x+1$ in $H_j$ is given by:

$$j + E_j \cdot e^{-\frac{x}{D_j \cdot \mathrm{N}}} \tag{4}$$

This might correspond to a model of GSAT's search in which there are a certain number of flips of size $j+1$ in each region $H_j$, and the probability of making a $j+1$ flip is merely dependent on the number of such flips left; the rest of the time, GSAT is obliged to make a flip of size $j$. Our data from 1000 tries fitted this model well, giving values of $R^2$ of 96.8% for $H_1$ and 97.5% for $H_2$. The regression gave estimates for the parameters of: $D_1 = 0.045$, $E_1 = 0.25$, $D_2 = 0.025$, $E_2 = 0.15$. Not surprisingly, since the region $H_3$ is very short, data was too noisy to obtain a better fit with the model (4) than with one of linear decay. These results support our conjecture, but more experiments on larger problems are needed to lengthen the region $H_j$ for $j \geq 3$.

## 6. Theoretical Conjectures

Empirical results like those given in §5 can be used to direct efforts to analyse algorithms theoretically. For example, consider the plateau region of GSAT's search. If the model (1) applies also to successful tries, the asymptotic score is L, giving

$$S(x) \quad = \quad \mathrm{L} - C \cdot \mathrm{N} \cdot e^{-\frac{x}{A \cdot \mathrm{N}}}$$

Differentiating with respect to $x$ we get,

$$\frac{\mathrm{d}S(x)}{\mathrm{d}x} \quad = \quad \frac{C}{A} \cdot e^{-\frac{x}{a \cdot \mathrm{N}}} \quad = \quad \frac{\mathrm{L} - S(x)}{A \cdot \mathrm{N}}$$

The gradient is a good approximation for $D_x$, the average size of a flip at $x$. Hence,

$$D_x \quad = \quad \frac{\mathrm{L} - S(x)}{A \cdot \mathrm{N}}$$

Our experiments suggest that downward flips and those of more than +1 are very rare on the plateau. Thus, a good (first order) approximation for $D_x$ is as follows, where $prob(D_x = j)$ is the probability that a flip at $x$ is of size $j$.

$$D_x \quad = \quad \sum_{j=-\mathrm{L}}^{\mathrm{L}} j \cdot prob(D_x = j) \quad = \quad prob(D_x = 1)$$

Hence,

$$prob(D_x = 1) \quad = \quad \frac{\mathrm{L} - S(x)}{A \cdot \mathrm{N}}$$

That is, on the plateau the probability of making a flip of size +1 may be directly proportional to $L - S(x)$, the average number of clauses remaining unsatisfied and inversely proportional N, to the number of variables. A similar analysis and result can be given for $prob(D_x = j+1)$ in the hill-climbing region $H_j$, which would explain the model (4) proposed in §5.

If our theoretical conjecture is correct, it can be used to show that the mean number of flips on successful tries will be proportional to $N \ln N$. Further investigation, both experimental and theoretical, will be needed to determine the accuracy of this prediction. Our conjectures in this section should be seen as conjectures as to what a formal theory of GSAT's search might look like, and should be useful in determining results such as average runtime and the optimal setting for a parameter like Max-flips. In addition, if we can develop a model of GSAT's search in which $prob(D_x = j)$ is related to the number of unsatisfied clauses and N as in the above equation, then the experimentally observed exponential behaviour and linear scaling of the score will follow immediately.

## 7. Related Work

Prior to the introduction of GSAT in (Selman et al., 1992), a closely related set of procedures were studied by Gu (Gu, 1992). These procedures have a different control structure to GSAT which allows them, for instance, to make sideways moves when upwards moves are possible. This makes it difficult to compare our results directly. Nevertheless, we are confident that the approach taken here would apply equally well to these procedures, and that similar results could be expected. Another "greedy algorithm for satisfiability" has been analysed in (Koutsoupias & Papadimitriou, 1992), but our results are not directly applicable to it because, unlike GSAT, it disallows sideways flips.

In (Gent & Walsh, 1993) we describe an empirical study of GenSAT, a family of procedures related to GSAT. This study focuses on the importance of randomness, greediness and hill-climbing for the effectiveness of these procedures. In addition, we determine how performance depends on parameters like Max-tries and Max-flips. We showed also that certain variants of GenSAT could outperform GSAT on random problems. It would be very interesting to perform a similar analysis to that given here of these closely related procedures.

GSAT is closely related to simulated annealing (van Laarhoven & Aarts, 1987) and the Metropolis algorithm, which both use greedy local search with a randomised method of allowing non-optimal flips. Theoretical work on these algorithms has not applied to SAT problems, for example (Jerrum, 1992; Jerrum & Sorkin, 1993), while experimental studies of the relationship between GSAT and simulated annealing have as yet only reached tentative conclusions (Selman & Kautz, 1993b; Spears, 1993).

Procedures like GSAT have also been successfully applied to constraint satisfaction problems other than satisfiability. For example, (Minton, Johnston, Philips, & Laird, 1990) proposed a greedy local search procedure which performed well scheduling observations on the Hubble Space Telescope, and other constraint problems like the million-queens, and 3-colourability. It would be very interesting to see how the results given here map across to these new problem domains.

## 8. Conclusions

We have described an empirical study of search in GSAT, an approximation procedure for satisfiability. We performed detailed analysis of the two basic phases of GSAT's search, an initial period of fast hill-climbing followed by a longer period of plateau search. We have shown that the hill-climbing phases can be broken down further into a number of distinct phases each corresponding to progressively slower climbing, and each phase lasting twice as long as the last. We have also shown that, in certain well defined problem classes, the average behaviour of certain important features of GSAT's search (the average score and the average branching rate at a given point) scale in a remarkably simple way with the problem size We have also demonstrated that the behaviour of these features can be modelled very well by simple exponential decay, both in the plateau and in the hill-climbing phase. Finally, we used our experiments to conjecture various properties (*eg.* the probability of making a flip of a certain size) that will be useful in a theoretical analysis of GSAT. These results illustrate how carefully performed experiments can be used to guide theory, and how computers have an increasingly important rôle to play in the analysis of algorithms.

## Acknowledgements

This research is supported by a SERC Postdoctoral Fellowship to the first author and a HCM Postdoctoral fellowship to the second. We thank Alan Bundy, Ian Green, and the members of the Mathematical Reasoning Group for their constructive comments and for the quadrillion CPU cycles donated to these and other experiments from SERC grant GR/H/23610. We also thank Andrew Bremner, Judith Underwood, and the reviewers of this journal for other help.